\documentclass{article}

\usepackage{microtype}
\usepackage{graphicx}
\usepackage{amsmath}
\usepackage{subfigure}
\usepackage{booktabs} 
\usepackage{tabularx}
\usepackage{multirow}
\usepackage{xcolor}
\usepackage{hyperref}

\usepackage[accepted]{icml2024}

\usepackage{amsmath}
\usepackage{amssymb}
\usepackage{mathtools}
\usepackage{amsthm}
\usepackage{lipsum}
\usepackage[capitalize,noabbrev]{cleveref}

\theoremstyle{plain}

\theoremstyle{definition}

\theoremstyle{remark}

\usepackage{hyperref}

\icmltitlerunning{Layer-wise LoRA fine-tuning: a similarity metric approach}

\begin{document}

\twocolumn[
\icmltitle{Layer-wise LoRA fine-tuning: a similarity metric approach}

\icmlsetsymbol{equal}{*}

\begin{icmlauthorlist}
\icmlauthor{Keith Ando Ogawa}{usp}
\icmlauthor{Bruno Lopes Yamamoto}{usp}
\icmlauthor{Lucas Lauton de Alcantara}{usp}
\icmlauthor{Lucas Pellicer}{icti}
\icmlauthor{Rosimeire Pereira Costa}{icti}
\icmlauthor{Edson Bollis}{icti}
\icmlauthor{Anna Helena Reali Costa}{usp}
\icmlauthor{Artur Jordao}{usp}
\end{icmlauthorlist}

\icmlaffiliation{usp}{Escola Politécnica, Universidade de São Paulo, São Paulo, Brazil}
\icmlaffiliation{icti}{Instituto de Ciência e Tecnologia Itaú (ICTi), São Paulo, Brazil}

\icmlkeywords{Machine Learning, PEFT, LoRA}

\vskip 0.3in
]

\printAffiliationsAndNotice{}

\begin{abstract}
Pre-training Large Language Models (LLMs) on web-scale datasets becomes fundamental for advancing general-purpose AI. In contrast, enhancing their predictive performance on downstream tasks typically involves adapting their knowledge through fine-tuning. Parameter-efficient fine-tuning techniques, such as Low-Rank Adaptation (LoRA), aim to reduce the computational cost of this process by freezing the pre-trained model and updating a smaller number of parameters. In comparison to full fine-tuning, these methods achieve over 99\% reduction in trainable parameter count, depending on the configuration. Unfortunately, such a reduction may prove insufficient as LLMs continue to grow in scale.
In this work, we address the previous problem by systematically selecting only a few layers to fine-tune using LoRA or its variants. We argue that not all layers contribute equally to the model adaptation. Leveraging this, we identify the most relevant layers to fine-tune by measuring their contribution to changes in internal representations. Our method is orthogonal to and readily compatible with existing low-rank adaptation techniques.
We reduce the trainable parameters in LoRA-based techniques by up to 50\%, while maintaining the predictive performance across different models and tasks. Specifically, on encoder-only architectures, this reduction in trainable parameters leads to a negligible predictive performance drop on the GLUE benchmark. On decoder-only architectures, we achieve a small drop or even improvements in the predictive performance on mathematical problem-solving capabilities and coding tasks. Finally, this effectiveness extends to multimodal models, for which we also observe competitive results relative to fine-tuning with LoRA modules in all layers. Code is available at \href[]{https://github.com/c2d-usp/Layer-wise-LoRA-with-CKA}{GitHub}
\end{abstract}

\section{Introduction}

Pre-trained Large Language Models (LLMs) exhibit remarkable performance on natural language processing and now shape a new wave of progress in solving complex reasoning tasks~\cite{Bengio:2025,AiIndex:2025}.
However, this effectiveness comes with a substantial computational cost during the training and fine-tuning phases of LLMs.
For example, a regular 16-bit fine-tuning of LLaMA with 65B parameters requires over 780GB of GPU memory~\cite{qlora}. Such computational and infrastructural demands drive the industry to lead the most significant advances in frontier AI research, as academia and small research centers lack resources to develop large and high-capacity models~\cite{AiIndex:2025,Morrison:2025}.

To address the previous issues, parameter-efficient fine-tuning (PEFT) methods emerge as alternatives to enable fine-tuning in low-resource environments~\cite{peft-survey}. These techniques reduce the number of trainable parameters by freezing the majority of them and keeping, or introducing, a small amount to optimize~\cite{peft-survey}. In particular, Low-Rank Adaptation (LoRA) gains prominence due to its simplicity and effectiveness when fine-tuning and no additional inference latency~\cite{lora}. 
Building upon the hypothesis that the change in weights during model adaptation has low intrinsic rank, LoRA operates by freezing the original weights of the model and adapting low-rank decomposition matrices, which leads to a substantial reduction of the adjustable parameters~\cite{lora}.
Formally, instead of optimizing the entire pre-trained weight matrix \( W_0 \in \mathbf{R}^{d \times k} \) as in full fine-tuning, LoRA adapts two matrices \( B \in \mathbf{R}^{d \times r} \) and \( A \in \mathbf{R}^{r \times k} \), with \( r \ll \min(d,k) \). Their product is the update $\Delta W \in \mathbf{R}^{d \times k}$, further scaled by $\frac{\alpha}{r}$ and added to $W_0$ to generate the fine-tuned matrix $W \in \mathbf{R}^{d \times k}$ as follows:
\begin{equation}
	W = W_0 + \Delta W=W_0 + \frac{\alpha}{r}BA.
\end{equation}
\begin{figure*}[!htb]
	\centering
	\includegraphics[width=1.0\linewidth]{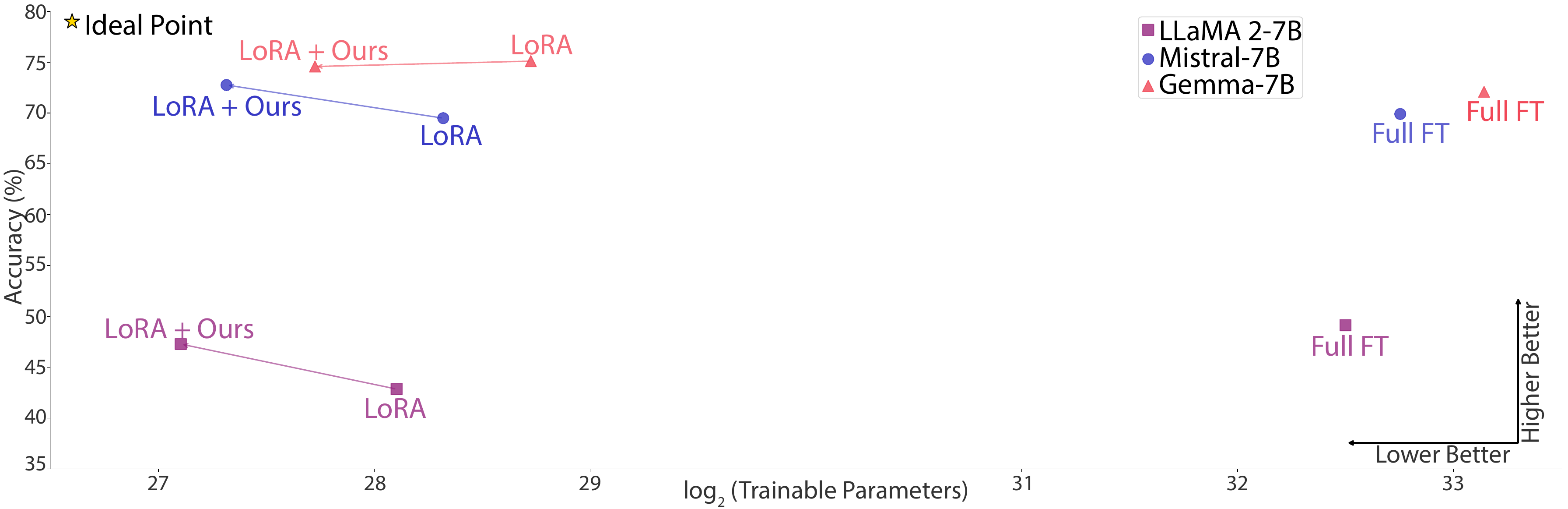}
	\caption{Trade-offs between LoRA with and without our method. We fine-tune LLaMA 2-7B, Mistral-7B and Gemma-7B on MetaMathQA and evaluate on GSM8K. Notably, our method reduces the number of trainable parameters while preserving predictive performance (in this case, increasing accuracy) compared to fine-tuning all layers with LoRA modules (standard practice). We observe the same behavior across different architectures, including multimodal models.}
	\label{fig:teaser}
\end{figure*}

\noindent 
Despite significantly reducing the number of adjustable parameters, the remaining ones can still pose a prohibitive computational burden.
Regardless of the existing challenges, LoRA and most of its variants approach the problem at the matrix level, focusing on aspects such as initialization strategies~\cite{pissa} and decomposition formats~\cite{adalora}, while more recent methods explore high-rank updates~\cite{hira}. Higher-level choices are typically not part of these methods, despite their potential to further reduce the computational cost. In particular, determining a selection of transformer layers to fine-tune is a promising approach that is orthogonal to these methods. Unless otherwise stated, in this work, fine-tune refers to LoRA fine-tune. In this direction, recent literature indicates that layers contribute unequally to changes in internal representations. More concretely, Skean et al.~\citeyear{layer-representation-study} refute the common assumption that final-layer representations are the most useful for downstream tasks, implying that some low and mid-depth layers play crucial roles in various tasks. Jin et al.~\citeyear{concept-depth} suggest that LLMs process tasks of different complexities in different layers, from which it follows that the contribution of a layer is task-specific. 
This context gives rise to a natural question: \emph{Is it possible to systematically choose the most important layers for a specific task adaptation?}

To answer this question, we propose a simple but effective method that explores the similarities between internal representations to identify the most important transformer layers of a model for a specific task. By measuring the importance of an entire transformer layer, our technique becomes orthogonal to LoRA-like methods, enabling easy integration to push the computational efficiency even further. Figure~\ref{fig:teaser} depicts this efficiency gain: we achieve results comparable to fine-tuning with all LoRA modules (standard practice) while using half of them.

\noindent \textbf{Research statement and contributions.} In general, our work has the following research statement. \emph{Given a LLM and a task, we can effectively choose a subset of transformer layers to fine-tune by measuring their participation in changes on internal representations. We quantify this relevance to the task by assessing the similarity between the input and output representations of a layer. The greater the contribution of a layer to the representation, the lower the similarity between its inputs and outputs. This subset comprises a configuration that, when fine-tuned, leads to an effective predictive ability while notably reducing the number of trainable parameters.}

Among our contributions, we highlight the following. We introduce a novel systematic method for choosing transformer layers to fine-tune that relies on layer contribution to changes in internal representations. From a practical perspective, we contribute to the progress of computational efficiency involving the fine-tuning of large language and multimodal models. Featuring a layer-level granularity, our technique enables a further reduction in the number of trainable parameters in constrained scenarios. From a theoretical perspective, our results suggest not only that certain layers are more influential on the fine-tuning process for a specific task, but also that similarity metrics are adequate tools to find them, overperforming naive strategies and existing methods.

Extensive experiments confirm our statement and contributions. Specifically, for encoder-only models, we achieve a 50\% reduction in backbone trainable parameters compared to fine-tuning all layers with LoRA, with a maximum drop in predictive performance of 2.5 percentage points in the average score of the GLUE benchmark. Moreover, on decoder-only models, we obtain the same reduction, but with improvements in predictive performance on most tasks for LLaMA 2-7B and Mistral-7B-v0.1, and minor drops for Gemma-7B. Figure~\ref{fig:teaser} illustrates these results and highlights the competitive predictive performance even with half the parameter count. Finally, for this computational budget, we observe similar results with the multimodal model $\text{LLaVA-1.5-7B}$.

\section{Related Work}
\noindent \textbf{LoRA and its variants.} As LLMs continue to grow in scale, PEFT techniques arise as promising solutions to enhance performance-to-parameter ratio in the fine-tuning process~\cite{peft-survey}.
Recent efforts in this research area concentrate on LoRA-based methods, as LoRA leverages a simple but effective approach: introducing low-rank matrix decompositions to adapt large models with significantly fewer trainable parameters.~\cite{lora}.
Extending LoRA, AdaLoRA employs an SVD-like decomposition and dynamically distributes the parameter budget among the matrices according to their importance score~\cite{adalora}. From a different perspective, PiSSA preserves the low-rank decomposition structure of LoRA, but initializes it with the principal singular values and vectors of the pre-trained matrix ($W_0$)~\cite{pissa}.

Typically, LoRA and most of its variants limit the rank of the adjustment matrix to the $r$ hyperparameter, constraining its representation expressiveness. On the other hand, HiRA guarantees high-rank updates by using the Hadamard product to build its adjustments~\cite{hira}. In contrast to these and other LoRA-based methods, our approach operates in an orthogonal direction, offering an additional option to further reduce the computational cost by selecting a set of transformer layers to fine-tune through a computationally efficient process. Our method works before starting the fine-tuning phase and relies on inference, avoiding expensive additional training steps or any operation that interferes with fine-tuning latency. We confirm that our technique works well not only with LoRA but also with its modern variants, such as PiSSA~\cite{pissa}.

\begin{figure}[!t]
	\centering
	\vspace{16pt}
	\includegraphics[width=1.0\columnwidth]{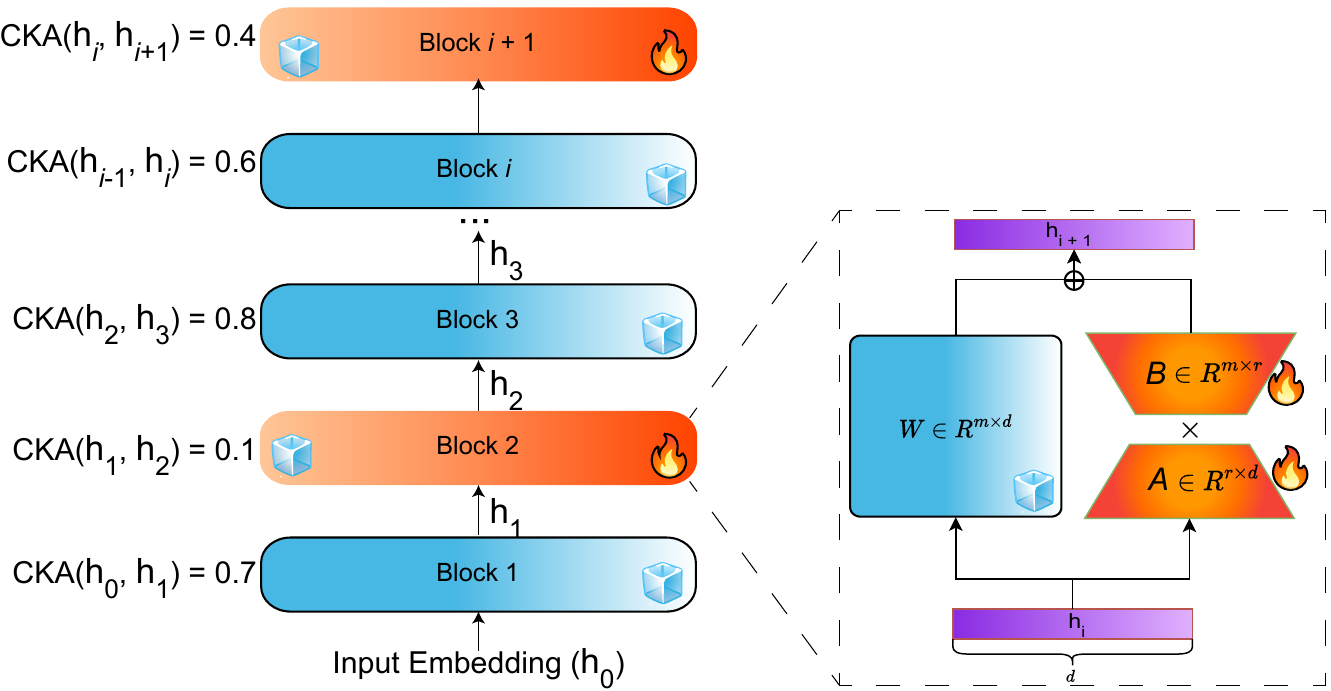}
	\caption{Overview of our method. We only fine-tune the subset of layers with the lowest similarity between their input and output representation (red parts).}
	\label{fig:teaset_method}
\end{figure}

\noindent \textbf{Internal Representations and Similarity Metrics.} The analysis of internal representations is a popular paradigm to better understand deep learning models~\cite{concept-depth,layer-representation-study}. Specifically, representation similarity metrics become powerful tools to explore complex phenomena~\cite{resi,cka}. As a concrete example, Jiang et al.~\citeyear{tracing-similarities} use similarity metrics to improve inference latency by exploring the phenomenon of \emph{saturation of events}, where model predictions are fully constructed at a specific layer and remain unchanged through the subsequent transformations. In the context of model compression, Pons et al.~\citeyear{effective-layer-pruning} employ similarity metrics as an estimation of layer importance in a layer pruning procedure.
Rather than promoting architecture changes, in our work, we adopt a similarity metric as an importance measure for selecting layers to fine-tune. Our method quantifies the relevance of a transformer layer to changes in internal representations by assessing the dissimilarity between input and output representations. We select layers with high dissimilarity to fine-tune, as we associate this with impact on internal representations, see Figure~\ref{fig:teaset_method}.

\noindent \textbf{Layer Selection.} In the vision domain, Lee et al.~\citeyear{surgical-finetuning} introduce the term \emph{surgical fine-tuning}: the practice of fine-tuning only a small contiguous subset of layers. Additionally, they propose a gradient-based metric to assign an importance value to a block of layers. Applying this idea to natural language processing, Lodha et al.~\citeyear{surgical-bert} establish Fisher Information Matrix (FIM) score as a measure of parameter importance and assess its efficacy on encoder-only models. Our work also addresses the selection of layers to fine-tune, however, we approach the problem through the lens of internal representations. Although our technique shares properties with previous methods, in the sense of being task-specific, we contribute to the domain by proposing a novel perspective, focusing on interactions with other PEFT methods and investigating its effectiveness on modern architectures. Furthermore, our method not only exhibits better predictive performance in comparison to the FIM approach, but also leverages a computationally cheaper process as it relies on inferences instead of expensive training steps.

\section{Preliminaries and Proposed Method}
\label{sec:prel}
\noindent \textbf{Problem statement.} Recent literature suggests that while some transformer layers play crucial roles in the construction of representations, others may be less important for specific models and data~\cite{tracing-similarities,concept-depth,layer-representation-study}. This evidence indicates that fine-tuning only the most important layers for a specific task could further improve the parameter efficiency of existing PEFT methods. Building upon this idea, we pose the following question: \emph{how to select a subset of transformer layers to finetune that matches the predictive performance of fine-tuning all layers?} Here, the term layer refers to the entire transformer block, composed of normalizations, skip connections, attention modules, and multi-layer perceptron modules.

\noindent \textbf{Definitions.} Let \( \mathcal{F}\) be a pre-trained large language model composed of \( M \) transformer layers, written as \( L_1, L_2, \dots, L_M \), where the embedding output is sequentially transformed from \( L_1 \) to \( L_M \). Define \( D = \{(x_j, y_j)\}_{j=1}^t \) as a downstream dataset of \( t \) samples for which we want to fine-tune \(\mathcal{F}\). For each layer \(L_i\), we define an internal representation \(R_i\) as the set of token representations at a specific position in the output of \(L_i\) across inputs $x_j$. Similarly, we define \(R_0\) as the embedding layer output representation.

Given  \( \mathcal{F}\) and \(D\), our layer importance metric assigns a value to each $L_i$ using $R_0, \dots, R_M$. From this importance, we can select the \( N\) most important ones, where \(N\) is a parameter, enabling us to control the computational cost. In our work, we explore the Centered Kernel Alignment (CKA) similarity metric to measure layer importance. According to Kornblith et al.~\citeyear{cka}, CKA is a normalized version of the Hilbert-Schmidt Independence Criterion (HSIC)~\cite{hsic}, which ensures desired properties for similarity metrics. For notational simplicity, assume $w$ and $z$ as two representations $R_i$ and $R_j$. Specifically, comparing $w$ and $z$, let $K_{ij} = k(\mathbf{z}_i, \mathbf{z}_j)$ and $Q_{ij} = q(\mathbf{w}_i, \mathbf{w}_j)$ where $k$ and $q$ are two kernels, \(p\) is the number of samples of the internal representation and  $H$ is the centering matrix $H_p = I_p - \frac{1}{p} \mathbf{1}\mathbf{1}^\text{T}$. The empirical estimator of HSIC is:
\begin{align}
	\text{HSIC}(K, Q) = \frac{1}{(p-1)^2}\text{tr}(KHQH).
	\label{eq:HSIC}
\end{align}

From Equation~\ref{eq:HSIC}, Kornblith et al.~\citeyear{cka} estimates CKA in terms of:
\begin{align}
	\text{CKA}(K, Q) = \frac{\text{HSIC}(K, Q)}{\sqrt{\text{HSIC}(K, K)\text{HSIC}(Q, Q)}}.
\end{align}

\noindent \textbf{Proposed Method.} Building upon the previous definitions, our technique proceeds as follows. 
Initially, we extract the internal representations from the output of each transformer layer and from the embedding layer output. In accordance with Klabunde et al.~\citeyear{resi}, the most representative token representations in the output of a LLM layer are the \textless CLS\textgreater \space token on encoder-only architectures and the last token on autoregressive architectures (decoder-only). As the final classifiers of encoder-only models use the \textless CLS\textgreater \space token and the decoder-only models use the last token to define the prediction of the inference, the authors argue that these tokens should contain all the relevant information.
Therefore, we use their representations as the layer output representation.

Subsequently, we use these representations to assign an importance value \( I_{L_i} \in [0,1]\) to each layer $L_i$. We define our layer importance metric as one minus the similarity (dissimilarity), measured with CKA, of $R_{i-1}$ and $R_i$:
\begin{align}
	I_{L_i} = 1 - \text{CKA}(R_{i-1},R_i).
\end{align}
Importantly, given that $L_i$ is the only structure between $R_{i-1}$ and $R_i$, it is intuitive to attribute the changes in $R_i$, quantified by their dissimilarity, to $L_i$. We hypothesize that the layer with the highest dissimilarity between its input and output representations has the most impact on the changes in internal representations. For this reason, it is a promising candidate to fine-tune. 
Building upon the previous idea, we choose $N$ layers to fine-tune by selecting the top $N$ layers with the highest dissimilarities. Using this subset, we reduce the computational cost of any LoRA-like method, such as PiSSA~\cite{pissa}, with minimal predictive ability degradation.
Extensive experiments support our idea, indicating that this importance metric built with CKA is suitable to choose good subsets of transformer layers to fine-tune.

\section{Experiments}
\textbf{Experimental Setup.} We conduct experiments on encoder-only transformer architectures with $\text{RoBERTa}_{\text{base}}$~\cite{roberta} and $\text{DeBERTa-v3}_\text{base}$~\cite{deberta}. On decoder-only transformer architectures, we use LLaMA 2-7B~\cite{llama2}, Mistral-7B-v0.1~\cite{mistral}, and Gemma-7B~\cite{gemma}. On a multimodal context, we evaluate our method on LLaVA-1.5-7B~\cite{llava1.5}. 
Our evaluation tasks for LLMs encompass both natural language understanding (NLU) and generation (NLG). To ensure a fair comparison with previous work, we follow the setup by Hu et al.~\citeyear{lora} and Zhang et al.~\citeyear{adalora} for $\text{RoBERTa}_{\text{base}}$ and $\text{DeBERTa-v3}_\text{base}$, respectively, on NLU. Therefore, we set the LoRA rank to 8 and the alpha to 16 for both models. On NLG, we adopt the setup by Meng et al.~\citeyear{pissa} for LLaMA 2-7B, Mistral-7B-v0.1, and Gemma-7B. Thus, we set the LoRA rank and alpha to 128 for these models. For the multimodal model LLaVA-1.5-7B, we employ an authorial setup to fine-tune for science question answering. We also set the LoRA rank and alpha to 128 for the multimodal model.

\begin{table}[!b]
	\centering
	\caption{Results of $\text{RoBERTa}_{\text{base}}$ and $\text{DeBERTa-v3}_\text{base}$ fine-tuned on the GLUE benchmark. We report Matthew's correlation for CoLA, overall (matched and mismatched) accuracy for MNLI, Pearson correlation for STSB, and accuracy for other tasks.  $\text{DeBERTa-v3}_\text{base}$ results for full fine-tuning and LoRA~\cite{lora} are from PiSSA~\cite{pissa} work.  $\text{RoBERTa}_{\text{base}}$ results for full fine-tuning and LoRA are from LoRA~\cite{lora} work. The remaining results are from our experiments. We report the average of three runs.}
	\resizebox{\columnwidth}{!}{ 
		\small
		\begin{tabular}{@{}lXccccccccccc@{}}
			\toprule
			\textbf{Model \& Method} & \textbf{Params} & \textbf{MNLI} & \textbf{SST2} & \textbf{MRPC} & \textbf{CoLA} & \textbf{QNLI} & \textbf{QQP} & \textbf{RTE} & \textbf{STSB} &\textbf{ALL}\\ 
			\midrule
			$\text{DeBERTa-v3}_\text{base}$ (Full FT)     & 184M    & 89.90           & 95.63          & 89.46          & 69.19          & 94.03          & 92.40 & 83.75            & 91.60         & 88.25       \\
			$\text{DeBERTa-v3}_\text{base}$ (LoRA)    &  1.33M  & 90.65           & 94.95          & 89.95          & 69.82          & 93.87          & 91.99          & 85.20            & 91.60         & 88.50       \\
			$\text{DeBERTa-v3}_\text{base}$ (FIM + LoRA)      &   0.66M & 86.85  & 91.44 & 78.43 & 61.11 & 92.32          & 90.69          & 58.24   & 85.46& 80.82\\
			$\text{DeBERTa-v3}_\text{base}$ (Ours + LoRA)      &   0.66M & 89.93  & 94.84 & 89.38 & 67.16 & 93.80          & 91.93          & 87.84   & 90.90&88.23\\
			\midrule
			$\text{RoBERTa}_{\text{base}}$ (Full FT)     & 125M    & 87.60           & 94.80          & 90.20          & 63.60          & 92.80          & 91.90 & 78.70            & 91.20         & 86.40       \\
			$\text{RoBERTa}_{\text{base}}$ $(\text{LoRA})$ & 0.3M & 87.50 & 95.10 & 89.70 & 63.40 & 93.30 & 90.80 & 86.60 & 91.50 & 87.2 \\
			$\text{RoBERTa}_\text{base}$ $(\text{FIM + LoRA})$ & 0.15M & 86.90 & 93.65 & 89.22 & 58.73 & 93.01 & 89.67 & 71.60 & 90.45 & 84.16 \\
			$\text{RoBERTa}_\text{base}$ $(\text{Ours + LoRA})$ & 0.15M & 86.70 & 93.60 & 88.60 & 61.30 & 92.80 & 89.70 & 74.50 & 90.20 & 84.70 \\
			\bottomrule
			
		\end{tabular}
	}
	
	\label{table:NLU}
\end{table}

\noindent
\textbf{Evaluating on NLU.} For NLU, we assess our method with $\text{RoBERTa}_{\text{base}}$ and $\text{DeBERTa-v3}_\text{base}$ on the GLUE benchmark~\cite{glue}. For both models, we select six layers (half of the layers composing the architecture) to fine-tune with LoRA. In particular, we avoid considering the first layer for encoder-only models, see Appendix for more details. Table~\ref{table:NLU} summarizes the results.

We observe that applying our method and fine-tuning the subset of layers with LoRA results in a marginal predictive performance drop and a 50\% reduction in the backbone trainable parameters. As our method exhibits a comparable average predictive performance with a significant decrease in the number of trainable parameters, these results translate into a more effective use of computational resources. To the best of our knowledge, the work by Lodha et al.~\citeyear{surgical-bert} is the only existing approach to tackle the problem of selecting layers to fine-tune in transformer architectures. In contrast to our method, their technique involves a higher computational cost, as it requires calculating the FIM score over training steps, and also exhibits a lower predictive performance. Specifically, in comparison to the regular LoRA, we achieve the parameter reduction of 50\% with only an average drop in predictive performance of 0.27 percentage points with $\text{DeBERTa-v3}_\text{base}$. In contrast, the method by Lodha et al.~\citeyear{surgical-bert} suffers from an average drop of 7.68 with the same configuration. With $\text{RoBERTa}_{\text{base}}$, although our method also presents a greater average predictive performance compared to the FIM score approach~\cite{surgical-bert}, the gap is tight. We believe that this difference in results between models is due to the large amount of parameters outside of the layers in $\text{RoBERTa}_{\text{base}}$, as it possibly reduces the impact of layer selection for fine-tuning. The previous results support our research statement. In particular, we identify subsets of layers to fine-tune that exhibit competitive predictive performance while still significantly reducing the number of trainable parameters. Moreover, our method outperforms the existing approach, reinforcing its effectiveness.

\begin{table}[!b]
	\centering
	\small
	\caption{Results of LLaMA 2-7B, Mistral-7B-v0.1, and Gemma-7B. We report pass@1 on base tests for HumanEval and MBPP. For GSM8K and MATH, we report accuracy, as it is the standard metric. Results for full finetuning, LoRA~\cite{lora} and PiSSA are from PiSSA~\cite{pissa} work. The remaining results are from our experiments. We report the average of three runs with standard deviations.}
	\resizebox{\columnwidth}{!}{ 
		\begin{tabular}{ccccccc}
			\toprule
			\textbf{Model}&\textbf{Params}&\textbf{Strategy}&\textbf{GSM8K}&\textbf{MATH}&\textbf{HumanEval}&\textbf{MBPP}\\
			\midrule
			\multirow{5}{*}{LLaMA 2-7B}& 6738M & Full FT&49.13±0.21&7.29±0.22& 21.20±0.30& 35.59±0.25\\
			& 320M & LoRA&42.85±0.12&5.50±0.33& 18.35±0.31 & 35.50±0.14 \\
			& 320M & PiSSA&53.22±0.55&7.47±0.34& 21.92±0.38 & 37.24±0.63\\ 
			& 159M & LoRA + Ours&47.28±0.44&6.45±0.06& 21.77±0.97 & 37.67±0.40\\ 
			& 159M & PiSSA + Ours&\textbf{54.79±0.92}&\textbf{8.83±0.13}& \textbf{25.63±2.80} & \textbf{40.50±1.14}\\ 
			\midrule
			\multirow{5}{*}{Mistral-7B}  & 7240M & Full FT&69.91±0.25&18.64±0.35& 45.31±0.14&51.46±0.13 \\
			& 335M & LoRA&69.50±0.42&20.08±0.20& 43.78±0.34& 58.46±0.37 \\
			& 335M & PiSSA&\textbf{73.31±0.23}&\textbf{23.12±0.52}  & 46.88±0.25 & \textbf{62.55±0.58}\\ 
			& 167M & LoRA + Ours&72.76±0.88&20.85±0.75& 48.20±1.20 & 61.80±0.66\\ 
			& 167M & PiSSA + Ours&72.66±0.46&20.77±0.28& \textbf{49.40±0.60} & 60.90±0.17\\ 
			\midrule
			\multirow{5}{*}{Gemma-7B}& 8540M & Full FT&72.09±0.32&22.71±0.34& 47.02±0.27& 55.67±0.50\\
			& 400M & LoRA&75.11±0.64&30.41±0.48&	53.70±0.23&	65.58±0.29\\
			& 400M & PiSSA&\textbf{77.78±0.32}&\textbf{31.33±0.33}&\textbf{54.31±0.28}&\textbf{66.17±0.43}\\ 
			& 200M & LoRA + Ours&74.58±0.38&28.59±0.38& 51.56±2.33 & 62.07±1.94\\ 
			& 200M & PiSSA + Ours&74.83±1.07&27.75±0.32& 52.00±0.35 & 62.97±1.52\\ 
			\bottomrule
		\end{tabular}
	}
	\label{table:NLG}
\end{table}

\noindent
\textbf{Evaluating on NLG.} For NLG, we evaluate our method with LLaMA 2-7B, Mistral-7B-v0.1, and Gemma-7B on mathematical problem-solving capabilities and coding proficiency. On mathematical problem-solving capabilities, we fine-tune the models on MetaMathQA~\cite{metamath} and evaluate on GSM8K~\cite{gsm8k} and MATH~\cite{math} validation sets. On coding proficiency, we fine-tune the models on Code-Feedback~\cite{code-feedback} and evaluate on  HumanEval~\cite{humaneval} and MBPP~\cite{mbpp}. Once again, we select half of the total layers for all models. Table~\ref{table:NLG} introduces the results.

\begin{figure*}[!t]
	\centering
	\includegraphics[width=1.0\textwidth]{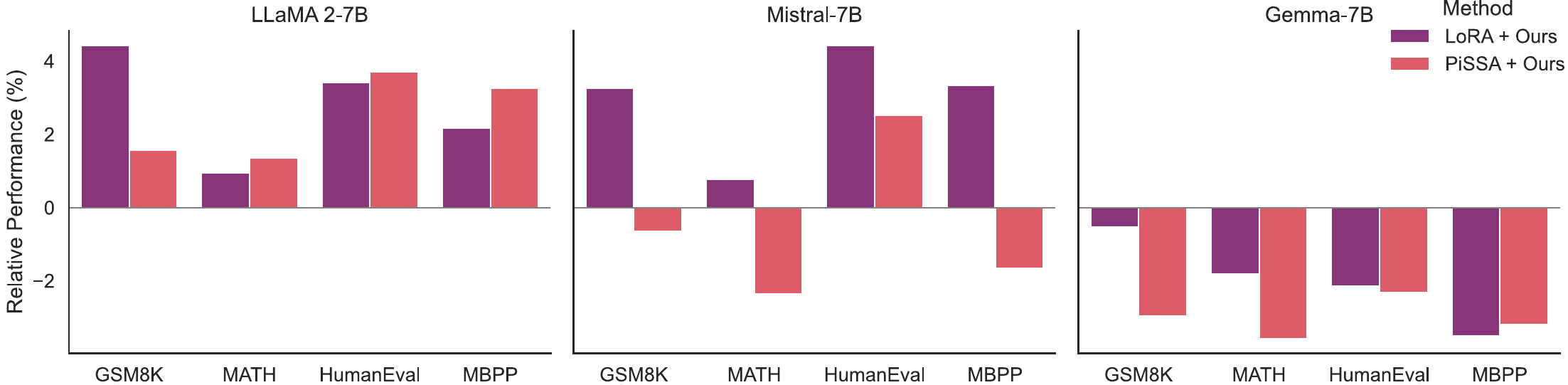}
	\caption{Relative performance of our method with different decoder-only models across multiple tasks. Each value is the difference between the predictive performance of a fine-tune in all layers and in a subset of 50\% of the layers; therefore, positive numbers indicate gains and negative numbers indicate drops.}	\label{fig:NLG_tasks}
\end{figure*}

In contrast to NLU, results across models share different patterns. Specifically, from Table~\ref{table:NLG}, we note that applying our method together with both LoRA and PiSSA on Gemma-7B leads to small predictive performance drops in all tasks. On the other hand, on LLaMA 2-7B and Mistral-7B-v0.1, even with a substantial reduction of trainable parameters, most tasks had an improvement in predictive performance. Figure~\ref{fig:NLG_tasks} illustrates these differences per model.

According to Figure~\ref{fig:NLG_tasks}, even though different decoder-only architectures exhibit distinct behaviours with our method, for each model, the results are homogeneous across tasks. Therefore, we believe that selecting a subset of layers to fine-tune may be model-specific. In addition, to the extent of our knowledge, it is unknown if every model has an ideal subset to fine-tune that leads to improvements or negligible degradation in predictive performance over fine-tuning all layers. However, in the following experiment, we show that our method produces subsets to fine-tune that outperform other simple strategies, indicating that $CKA$ dissimilarity identify adequate subsets.
Regardless of the diverse patterns across models, the results support our research statement. Specifically, we successfully fine-tune only a small subset of layers while matching or even surpassing the predictive performance of fine-tuning all layers.
Finally, we observe that applying our method with PiSSA also leads to effective predictive performances, reinforcing that our method is agnostic to the PEFT technique.

\begin{figure}[!b]
	\centering
	\includegraphics[width=1.0\columnwidth]{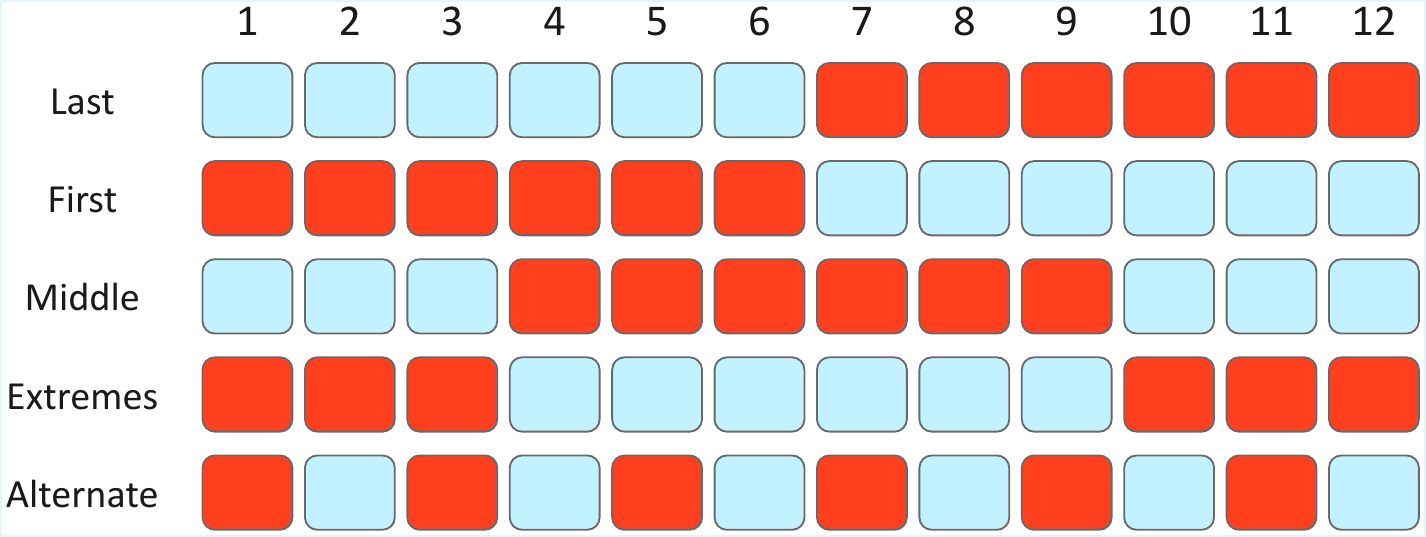}
	\caption{Simple layer selection strategies. Red rectangles represent trainable layers and blue rectangles frozen ones.}
	\label{fig:Naive_Heuristics}
\end{figure}

\noindent
\textbf{Is $CKA$ dissimilarity a good layer importance measure?} 
To evaluate the effectiveness of $CKA$ dissimilarity as a layer importance measure, we first replace it with naive heuristics to choose layers to fine-tune on the GLUE benchmark. We argue that a good layer importance measure should match or exceed the performance of simple heuristics, as one could use them instead of more complex criteria such as FIM~\cite{surgical-bert} or CKA. Figure~\ref{fig:Naive_Heuristics} introduces an overview of these heuristics.
Importantly, these heuristics remain fixed across different datasets, while our approach chooses subsets to fine-tune for each one.
Overall, our method outperforms naive strategies when selecting layers from $\text{RoBERTa}_{\text{base}}$ to fine-tune on the GLUE benchmark. We achieve significant improvements in comparison to most of them, while we made slight gains in comparison to selecting the middle layers. In particular, the differences on the GLUE average score between our approach and the \emph{Last}, \emph{First}, \emph{Middle}, \emph{Extremes}, and \emph{Alternate} heuristics are, in percentage points, respectively: 1.3, 2.5, 0.2, 1.5, and 0.4.
We hypothesize that selecting middle layers leads to competitive results due to the crucial role that they play in handling complex inputs~\cite{layer-representation-study}. Thus, aiming only at these structures is an easy way to guarantee a capacity to adapt to downstream tasks. To further investigate how our technique differentiates from this best-performing simple heuristic, we evaluate both methods on selecting layers to fine-tune LLaMA 2-7B, Mistral-7B-v0.1, and Gemma-7B on mathematical problem-solving capabilities and coding proficiency. Table~\ref{table:NLG_Middle} introduces the results.

Although the middle layers heuristic shows better predictive performance on Gemma-7B evaluations, our method exhibits overall better results. Specifically, while our technique maintains a maximum of 2.73 percentage points of difference compared to selecting the middle layers on its losing cases, the naive strategy reaches a catastrophic 7.57 percentage points gap with respect to the predictive performance of our method on HumanEval with Mistral-7B. We argue that our method avoids these catastrophic cases as it selects a subset of layers that strongly contribute to the changes in representations. In contrast, only looking at the same layers (e.g, middle ones) makes it possible to miss crucial layers for specific tasks. These results confirm the effectiveness of $CKA$ dissimilarity as a layer importance measure. Our method not only matches fine-tuning all layers, but also outperforms simple heuristics. Together, these observations suggest that we find effective subsets of layers to fine-tune, corroborating our research statement.

\begin{table}[!b]
	\centering
	\small
	\caption{Results of LLaMA 2-7B, Mistral-7B-v0.1, and Gemma-7B using our method and the middle layers heuristic to select layers. We report pass@1 on base tests for HumanEval and MBPP. For GSM8K and MATH, we report accuracy, as it is the standard metric. Bold values indicate the best results with LoRA and PiSSA of each model. We report the average of three runs with standard deviations.}
	\resizebox{\columnwidth}{!}{ 
		\begin{tabular}{ccccccc}
			\toprule
			\textbf{Model}&\textbf{Params}&\textbf{Strategy}&\textbf{GSM8K}&\textbf{MATH}&\textbf{HumanEval}&\textbf{MBPP}\\
			\midrule
			\multirow{3}{*}{LLaMA 2-7B}
			& 159M & LoRA + Ours&\textbf{47.28±0.44}&6.45±0.06& 21.77±0.97 & \textbf{37.67±0.40}\\ 
			
			& 159M & LoRA + Middle&45.33±0.87&\textbf{6.62±0.21}& \textbf{23.80±1.59} & 35.80±0.66\\ 
			
			& 159M & PiSSA + Ours&\textbf{54.79±0.92}&\textbf{8.83±0.13}& \textbf{25.63±2.80} & \textbf{40.50±1.14}\\ 
			& 159M & PiSSA + Middle&52.86±0.50& 8.44±0.08& 23.20±1.04 & 39.33±0.75\\ 
			
			\midrule
			\multirow{3}{*}{Mistral-7B}  
			& 167M & LoRA + Ours&\textbf{72.76±0.88}&\textbf{20.85±0.75}& \textbf{48.20±1.20} & 61.80±0.66\\ 
			& 167M & LoRA + Middle&71.95±0.53&20.69±0.55& 40.63±2.89 & \textbf{62.23±0.29}\\ 
			
			& 167M & PiSSA + Ours&72.66±0.46&20.77±0.28& \textbf{49.40±0.60} & 60.90±0.17\\ 
			& 167M & PiSSA + Middle&\textbf{72.80±0.57}&\textbf{21.51±0.08}& 46.53±2.48 & \textbf{61.30±0.82}\\ 
			
			\midrule
			\multirow{3}{*}{Gemma-7B}
			
			& 200M & LoRA + Ours&74.58±0.38&28.59±0.38&\textbf{51.56±2.33} & 62.07±1.94\\ 
			& 200M & LoRA + Middle&\textbf{76.72±0.53}&\textbf{29.03±0.22}& 50.20±1.25 & \textbf {64.80±0.50}\\ 
			& 200M & PiSSA + Ours&74.83±1.07& \textbf{27.75±0.32}& 52.00±0.35 & \textbf{62.97±1.52}\\ 
			& 200M & PiSSA + Middle&\textbf{76.52±0.59}&27.49±0.19& \textbf{52.63±1.44} & 62.60±1.91\\ 
			\bottomrule
		\end{tabular}
	}
	
	\label{table:NLG_Middle}
\end{table}

\noindent
\textbf{On the computational efficiency.} The absence of LoRA modules in a set of layers promotes a decrease in the computational cost of the fine-tuning process. To quantify this effect, we measure the training time and GPU memory usage in the fine-tune of  $\text{RoBERTa}_{\text{base}}$ on the GLUE benchmark and of $\text{LLaMA 2-7B}$ on MetaMathQA.

\begin{figure*}[tb]
	\centering
	\includegraphics[width=0.4960\linewidth]{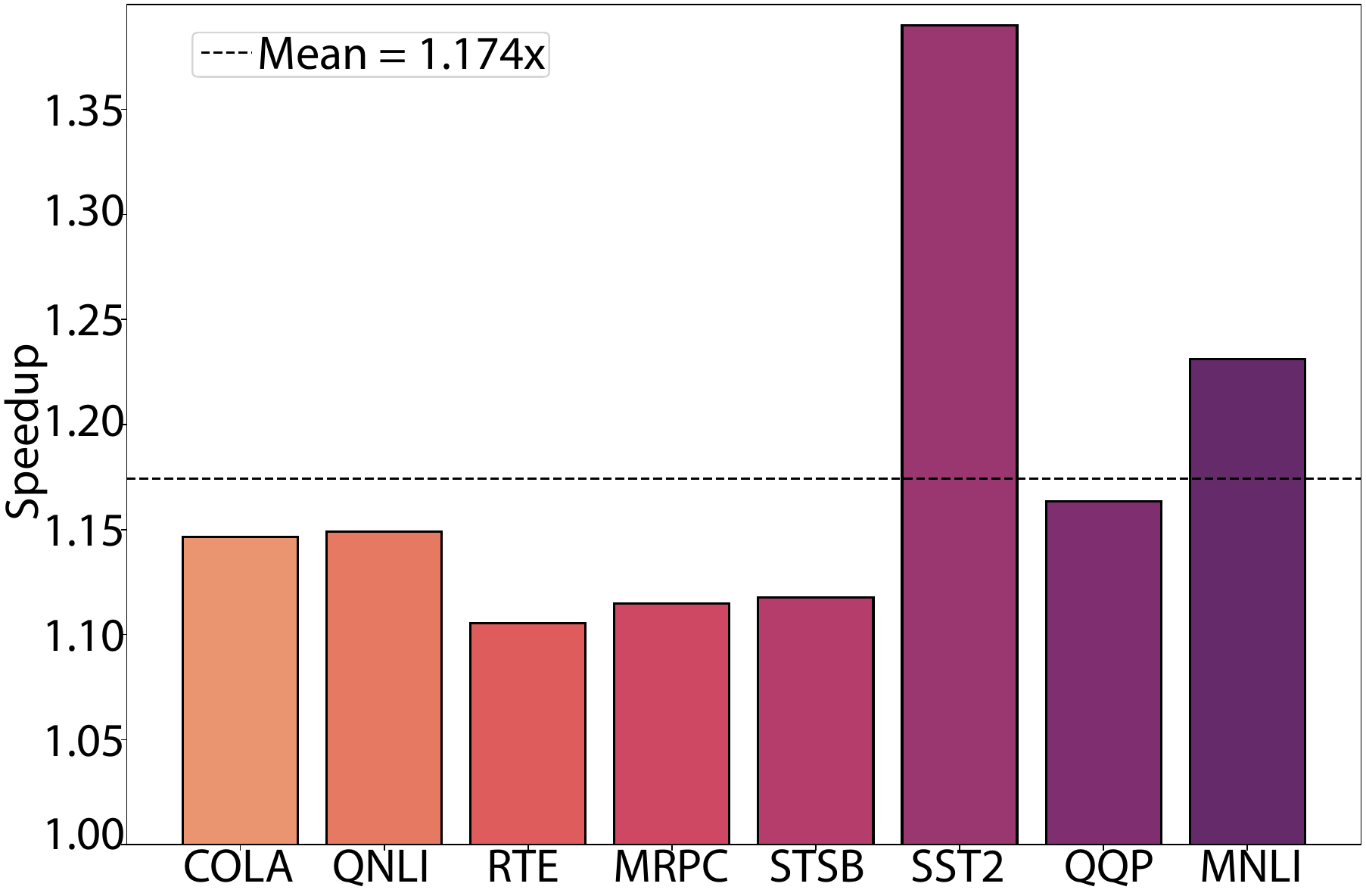}
	\includegraphics[width=0.4960\linewidth]{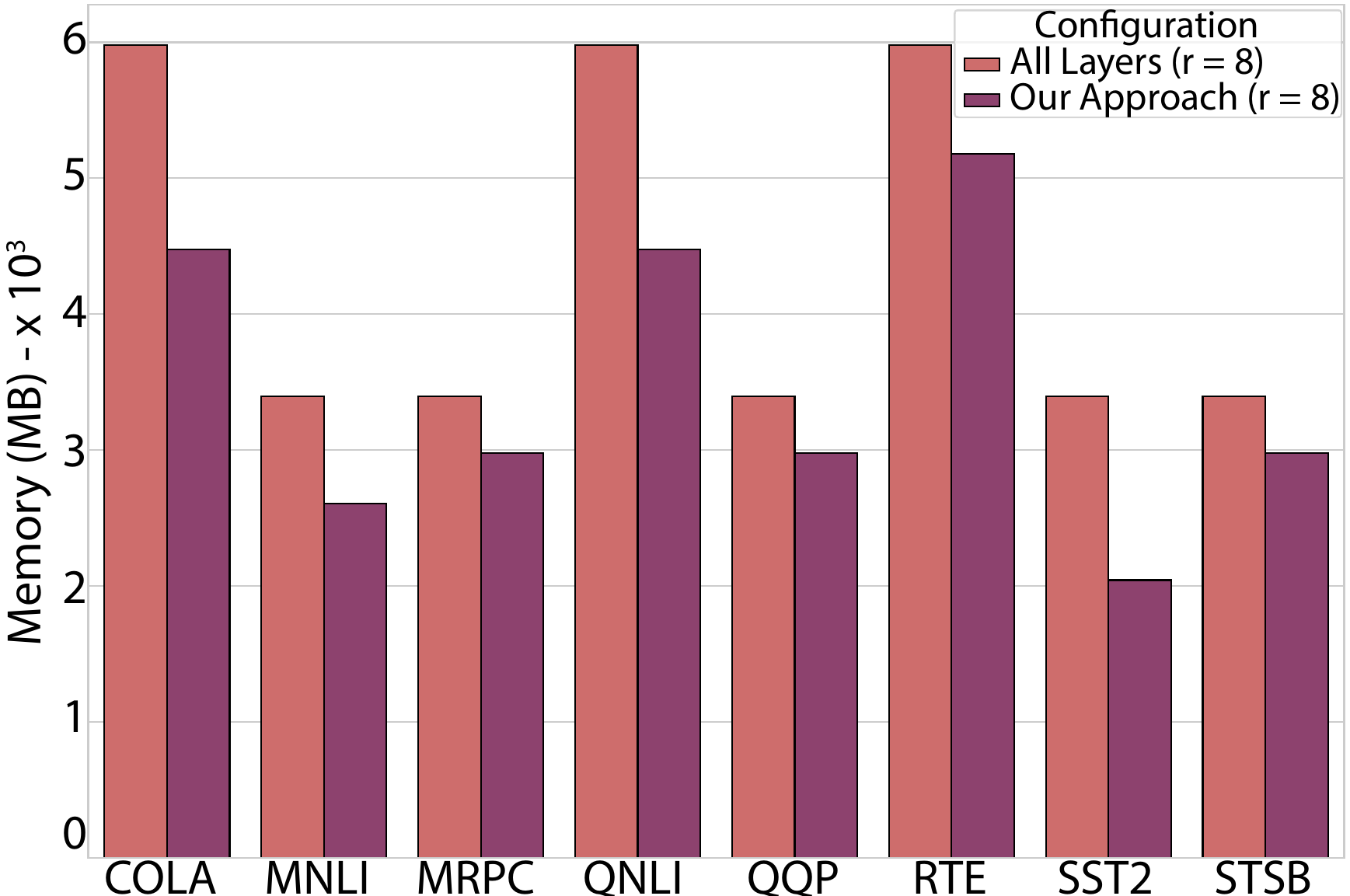}
	\caption{\textbf{Left: }Speedup in fine-tuning of $\text{RoBERTa}_{\text{base}}$ in the GLUE benchmark. We calculate the speedup in comparison to the fine-tuning with LoRA modules in all layers. \textbf{Right: }Mean memory allocated per task in fine-tuning of $\text{RoBERTa}_{\text{base}}$ in the GLUE benchmark. We measure the maximum memory usage for each training step.}
	\label{fig:eff}
\end{figure*}

Figure~\ref{fig:eff} (Left) shows that applying our method substantially reduces the training time with $\text{RoBERTa}_{\text{base}}$. The acceleration varies according to the selected layers. For example, while configurations that choose the first layers (e.g, RTE configuration) display an average acceleration of approximately 1.1, the ones that don't (e.g, SST2 configuration) exhibit higher values, reaching almost 1.4. We believe that this difference occurs because configurations with only the last layers can completely exclude the first ones from the backpropagation process. In contrast, others that include these layers need to calculate the gradient until it reaches them. Besides this additional speedup, the selection of layers to fine-tune yields significant acceleration. In LLaMA-2-7B fine-tune, we obtain a speedup of approximately 1.25 with a configuration that includes initial layers. This result, together with the average acceleration of 1.174 with $\text{RoBERTa}_{\text{base}}$, confirms the effective speedup our method provides across different scenarios.

Regarding memory usage, Figure~\ref{fig:eff} (Right) shows that the memory consumption shares the same pattern as acceleration: configurations concentrating deep layers leverage an additional reduction. Regardless of this behaviour,  $\text{RoBERTa}_{\text{base}}$ exhibits an average reduction in memory usage close to 20\%. Once again, LLaMA-2-7B presents a substantial efficiency gain by reducing the memory usage by around 15\% even when selecting the first layers. Therefore, we achieve a significant reduction in memory usage, even with unprivileged configurations.

\begin{table}[!b]
	\centering
	\caption{Results of selecting layer subsets of different sizes using our method with $\text{RoBERTa}_\text{base}$ on the GLUE benchmark. We report Matthew's correlation for CoLA, overall (matched and mismatched) accuracy for MNLI, Pearson correlation for STSB, and accuracy for other tasks. We report the average of three runs.}
	\resizebox{\columnwidth}{!}{ 
		\small
		\begin{tabular}{@{}lXccccccccccc@{}}
			\toprule
			\textbf{Model \& Method} & \textbf{Params} & \textbf{MNLI} & \textbf{SST2} & \textbf{MRPC} & \textbf{CoLA} & \textbf{QNLI} & \textbf{QQP} & \textbf{RTE} & \textbf{STSB} &\textbf{ALL}\\ 
			\midrule
			$\text{RoBERTa}_{\text{base}}$ (LoRA + Ours - One Layer)     & 0.025M    & 80.49           & 92.85          & 81.54          & 51.55          & 89.18          & 86.70 & 61.61            & 85.77         & 78.71       \\
			$\text{RoBERTa}_{\text{base}}$ (LoRA + Ours - Three Layers)      &	0.075M  &	85.18           &	\textbf{93.80}        &	87.17         &	59.93          &	91.98       &	88.18        &	73.40           &90.02          & 83.70       \\
			$\text{RoBERTa}_\text{base}$ $(\text{LoRA + Ours - Six Layers})$ & 0.15M & 86.74 & 93.60 & \textbf{88.60} & \textbf{61.30} & \textbf{92.80} & \textbf{89.70} & 74.50 & 90.17 & 84.66 \\
			$\text{RoBERTa}_{\text{base}}$ (LoRA + Ours - Nine Layers) & 0.22M & \textbf{87.03} & 90.34 & 86.93 & 60.82 & \textbf{92.80} & 89.54 & \textbf{77.61} & \textbf{90.52} & \textbf{84.96} \\
			\bottomrule
			
		\end{tabular}
	}
	\label{table:NLU_number_blocks}
\end{table}

\noindent
\textbf{On the number of selected layers.} For our previous experiments, we chose to select half of the total layers to fine-tune, promoting a parameter reduction of 50\%. Since this choice may be excessive or insufficient to fit a specific computational budget, an immediate question is: \emph{What is the behavior of our method with different numbers of selected layers?}

To answer this question, we evaluate $\text{RoBERTa}_\text{base}$ fine-tuned varying the number of layers (see parameter N in Section~\ref{sec:prel}) on the GLUE benchmark. Table~\ref{table:NLU_number_blocks} introduces the results and shows an interesting pattern: most tasks achieve their highest predictive performance or a value close to it with three or six selected layers. This result indicates that a few layers may be enough to address the majority of an adaptation to a downstream task. Regardless of this behavior, our technique exhibits a low predictive performance drop with different numbers of selected layers.
Therefore, we confirm that our method is suitable for various computational budgets, having the ability to handle different resource restrictions.

\begin{table}[!b]
	\centering
	\caption{Comparison between selecting a subset of layers with our method on $\text{RoBERTa}_\text{base}$. Following the GLUE benchmark standard practices, we report Matthew's correlation for CoLA, overall (matched and mismatched) accuracy for MNLI, Pearson correlation for STSB, and accuracy for other tasks. We report the average of three runs.}
	\resizebox{\columnwidth}{!}{ 
		\small
		
		\begin{tabular}{@{}lXccccccccccc@{}}
			\toprule
			\textbf{Model \& Method} & \textbf{Params} & \textbf{MNLI} & \textbf{SST2} & \textbf{MRPC} & \textbf{CoLA} & \textbf{QNLI} & \textbf{QQP} & \textbf{RTE} & \textbf{STSB} &\textbf{ALL}\\ 
			\midrule
			$\text{RoBERTa}_{\text{base}}$ $(\text{LoRA}_{r = 8})$ & 0.30M & 87.50 & 95.10 & 89.70 & 63.40 & 93.30 & 90.80 & 86.60 & 91.50 & 87.2 \\
			$\text{RoBERTa}_\text{base}$ $(\text{LoRA + Ours}_{r = 8})$ & 0.15M & 86.70 & 93.60 & 88.60 & 61.30 & 92.80 & 89.70 & 74.50 & 90.20 & 84.70 \\
			$\text{RoBERTa}_\text{base}$ $(\text{LoRA + Ours}_{r = 16})$ & 0.30M & 86.92 & 93.84 & 88.23 & 60.14 & 92.69 & 90.04 & 76.30 & 90.18 & 84.79 \\
			\bottomrule
			
		\end{tabular}
	}
	\label{table:NLU_rank16}
\end{table}

\noindent
\textbf{On the role of LoRA rank.}  The rank hyperparameter (r) of LoRA both defines the maximum rank of the adjustment and the number of parameters in the modules. Huang et al.~\citeyear{hira} associate an increase in rank with additional expressiveness in updates and predictive performance gains, highlighting the importance of this hyperparameter. As we reduce the number of parameters by using fewer modules, a natural question arises: \emph{Does redistributing all parameters to the most important layers, via a rank increase in their LoRA modules, improve predictive performance compared to fine-tuning all layers?}

To answer this question, we use our method to select a subset of layers to fine-tune $\text{RoBERTa}_\text{base}$ with LoRA, doubling the rank of the baseline setup to 16, on the GLUE benchmark. An important observation is that this setup maintains the same number of parameters as the baseline, as the selection of layers and the rank increase compensate for each other in terms of parameter count. When increasing the rank, we ensure that the ratio between rank and alpha hyperparameters of LoRA remains the same, as suggested by previous work~\cite{hira,dora,pissa}. Table~\ref{table:NLU_rank16} introduces the results and shows that a redistribution of parameters does not effectively improve task performance. We found these results counterintuitive, as recent work underscores the importance of the rank~\cite{hira}. However, the unique setup with rank increases in specific layers could explain this difference. 
Specifically, results on MRPC, CoLA, QNLI, and STSB exhibit a slight drop, while other tasks display minor gains. Overall, these changes translate into a 0.09 percentage points gain on average in comparison with our results with rank equal to eight.
We believe that the reason for this unexpected outcome is the nature of our approach. As we use the dissimilarity of representations, including multiple samples, we select layers that interact with general characteristics of the data. Our hypothesis is that low-rank adjustments are enough to address these coarse-grained patterns as they tend to be less complex in comparison to fine-grained details. Therefore, we argue that rank increase is ineffective with our method because we aim at interactions between layers and data that probably lack the necessity of additional expressiveness.

\noindent
\textbf{Effectiveness with multimodal models.} Beyond the language domain, we assess our method with the multimodal model LLaVA-1.5-7B~\cite{llava1.5} on the question answering benchmark ScienceQA~\cite{scienceqa}. We select half of the total layers of the language tower to fine-tune and evaluate with the LMMS-eval framework~\cite{lmms-eval}.

In this setting, our method achieves an average accuracy of 79.26\%, while fine-tuning with all LoRA modules exhibits an average accuracy of 79.49\%. This marginal difference of 0.23 percentage points, even with a reduction from 160M to 80M trainable parameters, suggests that our technique also chooses adequate subsets of layers for multimodal models. Therefore, we confirm the efficacy of our method on a multimodal model that has a different training process and architecture in comparison to the extensively addressed before LLMs.

\section{Conclusions}

LoRA and its variants demonstrate the capacity to drastically reduce the number of trainable parameters while maintaining the predictive performance of the fine-tuned model. In this work, we explore, through the lens of representation similarity, an orthogonal direction to these methods to further increase computational efficiency: the selection of layers to fine-tune with LoRA. Throughout extensive experiments, we show that our method successfully reduces the number of trainable parameters while preserving predictive performance in comparison to the regular LoRA. In particular, we observe predictive performance gains for LLaMA-2-7B and Mistral-7B with a 50\% reduction in trainable parameters, and negligible drops for other popular language models. Notably, we also achieve a competitive predictive performance with the multimodal model LLaVA-1.5-7B, demonstrating the effectiveness of our method beyond the language domain.  Finally, to the best of our knowledge, our work is the first to explore effective subsets of layers to fine-tune with LoRA across a wide set of models and tasks, including a multimodal architecture.

\noindent \textbf{Open questions and room for improvement} Despite the positive results, some questions remain unaddressed in this paper: 1) $I_{L_i}$ is an importance metric that measures the individual contribution of each layer. Is there an importance metric that successfully captures not only the contribution of a layer to the changes of internal representations, but also the interactions between them?  2) Instead of selecting the important layers to fine-tune, is it possible to define an ideal LoRA rank for each layer? 3) We select layers to fine-tune through a similarity metric perspective. Are there other effective approaches?
Although we do not address these questions in our study, we highlight them as promising research directions for future work.
\section{Acknowledgments}

The authors would like to thank Instituto de Ciência e Tecnologia Itaú (ICTi) and Programa de Bolsas Itaú (PBI). This study was financed in part by the Coordenação de Aperfeiçoamento de Pessoal de Nível Superior – Brasil (CAPES) – Finance Code 001. Anna H. Reali Costa would like to thank grant \#312360/2023-1 CNPq. 
This work was partially supported by the Instituto Nacional de Ciência e Tecnologia em Inteligência Artificial Responsável para Linguística Computacional, Tratamento e Disseminação de Informação (INCT-TILDIAR - CNPq grant \#408490/2024-1). 

\bibliography{refs}
\bibliographystyle{icml2024}

\newpage
\appendix
\section{The first layer of encoder-only models}
\label{sec:appendix_encoder}

\begin{figure}[!h]
	\centering
	\includegraphics[width=1.0\linewidth]{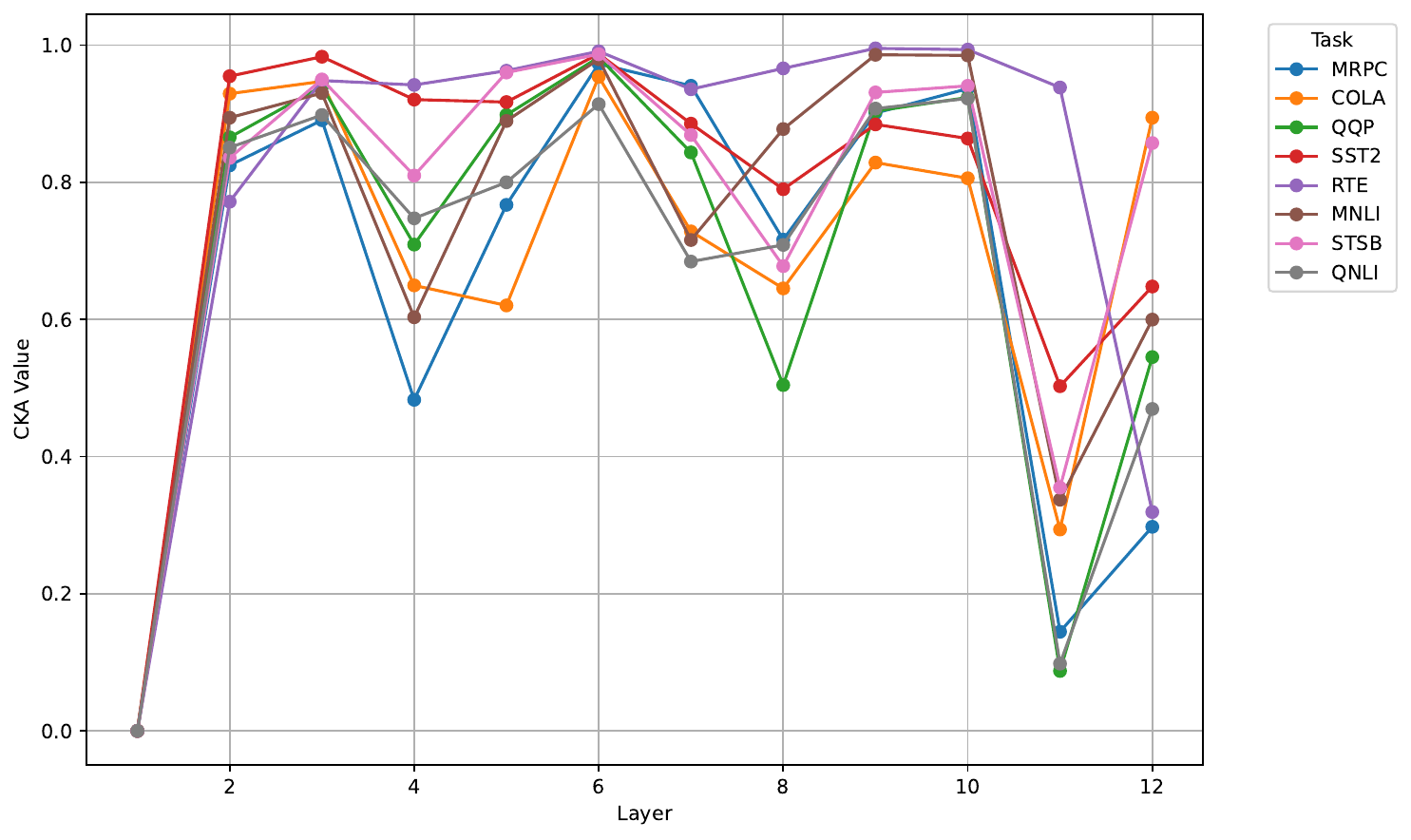}
	\caption{$CKA$ value per layer of $\text{RoBERTa}_{\text{base}}$ in different tasks of the GLUE benchmark.}
	\label{fig:cka_per_layer}
\end{figure}

We observe that our method assigns a counterintuitive importance value to the first layer of encoder-only models. Figure~\ref{fig:cka_per_layer} shows that the first layer input and output representations consistently display a near-zero similarity, which implies an $I_{L_1}$ close to its maximum value.
This phenomenon occurs because $\text{CKA}$ uses centered matrices in its computation. In the input of the first layer, the representation is still untouched. Therefore, as we take the $\text{<CLS>}$ token representation for every sample, the mean becomes this representation and the centered matrix zeroed, which causes the CKA to be zero in the first layer of encoder-only models.
Beyond this investigation, we conduct an experiment to evaluate the impact of the first layer on predictive performance. We pick a set of layers to fine-tune $\text{RoBERTa}_{\text{base}}$ in the GLUE benchmark with and without the first one. As we observe an average difference of less than 0.2 percentage points between these two configurations, we conclude that the first layer does not play a crucial role for all tasks. Therefore, to ensure consistency, we exclude the first layer from the set of candidates for our selected subset of layers.

\end{document}